\documentclass{llncs}

\usepackage{amssymb}
\usepackage{textcomp}
\usepackage{comment}
\usepackage{color}
\usepackage{graphicx}
\usepackage{amsmath}

\newcommand{\Beginproof}{{\em Proof.}  }
\newcommand{\Endproof}{\hfill$\Box$}

\begin{document}

\title{The Quantum Version Of Classification Decision Tree Constructing Algorithm C5.0}
\author{Kamil~Khadiev$^{1,2}$ \and Ilnaz Mannapov$^{2}$ \and Liliya Safina$^{2}$}

\institute{
Kazan Federal University, 18,  Kremlyovskaya st, Kazan, Russia, 420008
  \and
Kazan E. K. Zavoisky Physical-Technical Institute, 10/7,Sibirsky tract, Kazan, Russia, 420029
                \\ \email{kamilhadi@gmail.com, ilnaztatar5@gmail.com,  liliasafina94@gmail.com}
}

\maketitle

\begin{abstract}In the paper, we focus on complexity of  C5.0 algorithm for constructing decision tree classifier that is the models for the classification problem from  machine learning. In classical case the decision tree is constructed in $O(hd(NM+N \log N))$ running time, where  $M$ is a number of classes, $N$ is the size of a training data set, $d$ is a number of attributes of each element, $h$ is a tree height. Firstly, we improved the classical version, the running time of the new version is $O(h\cdot d\cdot N\log N)$. Secondly, we suggest a quantum version of this algorithm, which uses quantum subroutines like the amplitude amplification and the D{\"u}rr-H{\o}yer minimum search algorithms that are based on Grover's algorithm. The running time of the quantum algorithm is $O\big(h\cdot \sqrt{d}\log d \cdot N \log N\big)$ that is better than complexity of the classical algorithm. 
\end{abstract}


\section{Introduction}
\emph{Quantum computing} \cite{nc2010,a2017} is one of the hot topics in computer science of last decades.
There are many problems where quantum algorithms outperform the best known classical algorithms \cite{dw2001,quantumzoo,ks2019,kks2019}. 
%
Today quantum computing is often used in machine learning to speed up  construction of machine learning models or to predict a result for new input data \cite{aw2017,ko2018,ssp2015,ssp2014qnn}. Sometimes the learning process (construction) of a machine learning model takes a long time because of the large size of data. Even a small reduction in running time can provide a significant temporary benefit to the program. 

Decision trees are often used to build a classifier. Random forest\cite{GTB}, Gradient tree boosting\cite{fri} models are very popular and effective for solving classification and regression problems. These algorithms are based on decision trees. There are several algorithms for trees construction that are CART\cite{cart}, ID3\cite{ml2}, C4.5\cite{c45}, C5.0\cite{c50} and others. 
We consider C5.0 \cite{c50main} algorithm for decision tree classifiers. It works in $O(hd(NM+N \log N))$ running time, where $h$ is a height of a tree, $N$ is the size of a training set, $d$ is a number of attributes for one vector from the training set, and $M$ is a number of classes.

In this paper, firstly, we present an improved version of the classical algorithm that uses Self-balancing binary search tree \cite{cormen2001} and has $O(hdN \log N)$ running time.  As a self-balancing binary search tree we can use the AVL tree \cite{avl62,cormen2001} or the Red-Black tree \cite{g78,cormen2001}. Secondly, we describe a quantum version of the C5.0 algorithm. We call it QC5.0. The running time of QC5.0  is equal to $O\big(h\log d\sqrt{d}N \log N\big)$. The algorithm is based on generalizations of Grover's Search algorithm \cite{g96} that are amplitude amplification \cite{bhmt2002} and D{\"u}rr-H{\o}yer algorithm for minimum search \cite{dh96}.

The paper has the following structure. Section \ref{sec:prelims} contains preliminaries. Description of the classical version C4.5 and C5.0 algorithms are in Section \ref{sec:c50algo}. Section \ref{sec:improvement} contains improvements of classical algorithm. We provide the quantum algorithm in Section \ref{sec:qauntumalgo}.



\section{Preliminaries}\label{sec:prelims}
Machine learning \cite{ml1, ml2} allows us to predict a result using information about past events.  C5.0 algorithm is used to construct a decision tree for classification problem \cite{DT}.
Let us consider a classification problem in formal way. 

There are two sequences:  ${\cal X}=\{X^1, X^2, ..., X^N\}$ is a training data set and ${\cal Y}=\{y_1, y_2, ..., y_N\}$ is a set of corresponding classes. Here $X^i=\{x^i_1, x^i_2, ..., x^i_d\}$ is a vector of attributes, where $i\in\{1,\dots,N\}$, $d$ is a number of attributes, $N$ is a number of vectors in the training data set, $y_i \in C=\{1,\dots,M\}$ is a number of class of $X^i$ vector. An attribute $x^i_j$ is a real-valued variable or a discrete-valued variable, i.e. $x^i_j\in\{1,\dots,T_j\}$ for some integer $T_j$. Let $DOM_j=\mathbb{R}$ if $x_j$ is a real value; and $DOM_j=\{1,\dots,T_j\}$ if $x_j$ is a discrete-valued attribute. The problem is to construct a function $F:DOM_1\times\ldots\times DOM_d\to C$ that is called classifier. The function  classifies a new vector $X=(x_1,\dots,x_d)$ that is not from ${\cal X}$.

There are many algorithms to construct a classifier. Decision tree and the algorithm C5.0 for constructing a decision tree are a central subject of this work.

A decision tree is a tree such that each node tests some condition on input variables. Suppose $B$ is some test with outcomes ${b_1, b_2,\dots, b_t}$ that is tested in a node. Then, there are $t$ outgoing edges for the node for each outcome. Each leaf is associated with a result class from $C$. The testing process is the following. We start test conditions from the root node and go by edges according to a result of the condition. The label on the reached leaf is the result of the classification process. 

Our algorithm uses some quantum algorithms as a subroutine, and the rest part is classical. As quantum algorithms, we use query model algorithms. These algorithms can do a query to a black box that has access to the training data set and stored data. As a running time of an algorithm, we mean a number of queries to the black box. In a classical case, we use the classical analog of the computational model that is query model. 
We suggest \cite{nc2010} as a good book on quantum computing and \cite{a2017} for a description of the query model.




\section{The Observation of C4.5 and C5.0 Algorithms}\label{sec:c50algo}

We consider a classifier $F$ that is expressed by decision trees. 
This section is dedicated to the C5.0 algorithm for decision trees construction for the classification problem. This algorithm is the improved version of the algorithm C4.5, and it is the part of the commercial system See5/C5.0. C4.5 and C5.0 algorithms are proposed by Ross Quinlan\cite{DT}. Let us discuss these algorithms.
C4.5 belongs to a succession of decision tree learners that trace their origins back to the work of Hunt and others in the late 1950s and early 1960s \cite{h62}. Its immediate predecessors were ID3 \cite{q79}, a simple system consisting initially of about 600 lines of Pascal, and C4 \cite{q87}. 

\subsection{The Structure of the Tree}

Decision tree learners use a method known as divide and conquer to construct a
suitable tree from a training set ${\cal X}$ of vectors:
\begin{itemize}
    \item If all vectors in ${\cal X}$ belong to the same class $c\in C$, then the decision tree is a leaf labeled by $c$.
    \item Otherwise, let $B$ be some test with outcomes ${b_1, b_2,\dots, b_t}$ that produces a non-trivial partition of ${\cal X}$. Let ${\cal X}_i$ be the set of training vectors from ${\cal X}$ that has outcome $b_i$ of $B$. Then, the tree is presented in Figure \ref{fig:dt}. Here $T_i$ is a result of growing a decision tree for a set ${\cal X}_i$.
    \begin{figure}[h!]
    \centering
    \includegraphics[width=200px]{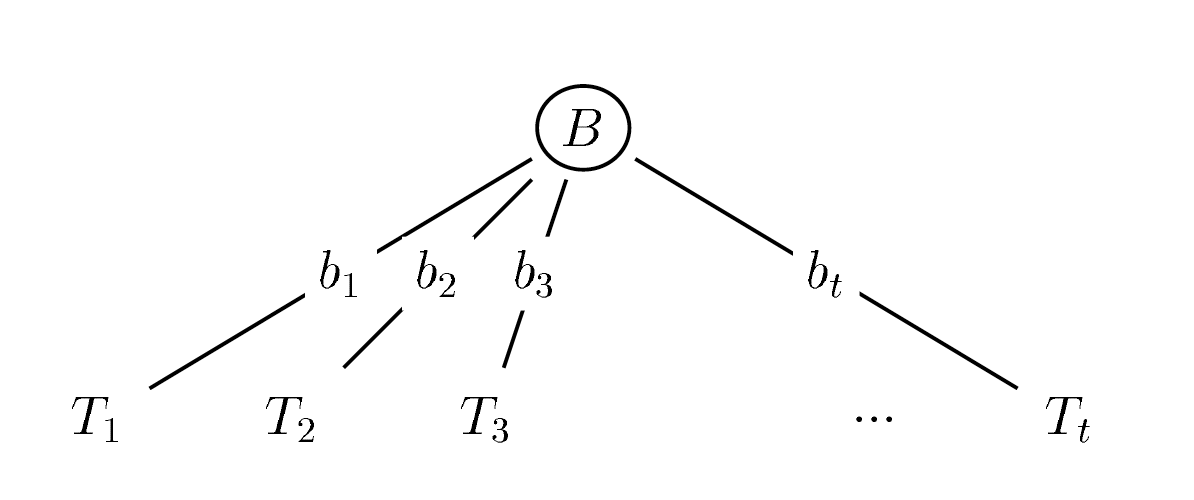}
    \caption{Testing $B$ with outcomes ${b_1, b_2,\dots, b_t}$. Here $T_i$ is a result of growing a decision tree for a set ${\cal X}_i$}\label{fig:dt}
    \end{figure}
\end{itemize}

C4.5 uses tests of three types, each of them involves only a single attribute $A_a$. Decision regions in the instance space are thus bounded by hyperplanes, each of them is orthogonal to one of the attribute axes.

\begin{itemize}
    \item If $x_j$ is a discrete-valued attribute from $\{1,\dots, T_j\}$, then possible tests are
    \begin{itemize}
        \item $x_j = ?$ with $T_j$ outcomes, one for each value from $\{1,\dots, T_j\}$. (This is the default test.)
        \item $x_j \in G$ where $G\subset \{1,\dots, T_j\}$. Tests of this kind are found by a greedy search that maximizes the value of the splitting criterion (It is discussed below).
    \end{itemize}
    \item If $x_j$ is a real-valued attribute, then a test is ``$x_j\leq\theta$'' with two outcomes that are ``true'' and ``false''. Here $\theta$ is a constant threshold. Possible values of $\theta$ are found by sorting the distinct values for $\{x^1_j,\dots, x^N_j\}$ set. Possible thresholds are values between each pair of adjacent values in the sorted sequence. So, if the training vectors from ${\cal X}$ have $d$ distinct values for $j$-th attribute, then $d-1$ thresholds are considered.
\end{itemize}

\subsection{Test Selection Procedure}

C4.5 relies on a greedy search, selecting a candidate test that maximizes a heuristic splitting criterion.

Two criteria are used in C4.5 that are information gain, and gain ratio.  Let $C_j=\{i:i\in\{1, \dots, |{\cal X}|\},y_i=j\}$ be a set of indexes of training vectors from ${\cal X}$ that belong to $j$-th class, for $j\in C=\{1,\dots,M\}$.  Let $RF(j;{\cal X})$ be a relative frequency of training vectors in $\cal X$ with indexes from $C_j$.
$RF(j; {\cal X}) = \frac{|C_j|}{|{\cal X}|}$
The information content of a message that identifies the class of vectors from ${\cal X}$ is 
$I({\cal X})=-\sum_{j=1}^{M} R F\left(j, {\cal X}\right) \log \left(R F\left(j, {\cal X}\right)\right)$
After that we split $X$ into subsets ${\cal X}_1, {\cal X}_2,\dots, {\cal X}_t$ with respect to a test $B$, the information gain is 

$G({\cal X}, B)=I({\cal X})-\sum_{i=1}^{t} \frac{\left|{\cal X}_{i}\right|}{|{\cal X}|} I\left({\cal X}_{i}\right).$
The potential information from the partition itself is

$P({\cal X}, B)=-\sum_{i=1}^{t} \frac{\left|{{\cal X}}_{i}\right|}{|{\cal X}|} \log \left(\frac{\left|{{\cal X}}_{i}\right|}{|{\cal X}|}\right)$.
The test $B$ is chosen such that it maximizes the gain ratio that is $\frac{G({\cal X};B)}{P({\cal X};B)}$.

\subsection{Notes on C5.0 algorithm}

C4.5 was superseded in 1997 by a commercial system See5/C5.0 (or C5.0 for short). The
changes encompass new capabilities as well as much-improved efficiency, and include the following items.
(1) A variant of boosting [24], which constructs an ensemble of classifiers that are later vote to give a final classification. Boosting often leads to a dramatic improvement in predictive accuracy.
    (2) New data types (e.g., dates), ``not applicable'' values, variable misclassification costs, and mechanisms for a pre-filtering of attributes.
    (3) An unordered rule sets that it is a situation when a vector is classified, all applicable rules are found and voted. This fact improves both the interpretability of rule sets and their predictive accuracy.
    (4) Greatly improved scalability of both decision trees and (particularly) rule sets (sets of if-then rules, representation of decision tree). Scalability is enhanced by multi-threading; C5.0 can take advantage of computers with multiple CPUs and/or cores.

More details are available in \cite{c50main} ( http://rulequest.com/see5-comparison.html).
At the same time, the process of choosing a test $B$ was not changed significantly. In this work, we focus on this process and improve its complexity.

\subsection{Running Time of the One-threading Tree Constructing Part of  C4.5 and C5.0 algorithms}

Let us remind the parameters of the model. $N$ is a number of vectors in a training set, $M$ is a number of classes, $d$ is a number of attributes (elements of vectors from the training set). 
Let the height of a constructing tree $h$ be a parameter of the algorithm. Let $RV$ be a set of indexes of real-valued attributes and let $DV$ be a set of indexes of discrete-valued attributes.

Let us describe the procedure step by step because we will use it for our improvements. Assume that we construct a binary tree of height $h$.

The main procedure is \textsc{ConstructClassifiers} that invoke a recursive procedure \textsc{FormTree} for constructing nodes. The main parameters of \textsc{FormTree} are $level$ that is an index of tree level; $tree$ that is a result subtree that the procedure will construct;  ${\cal X}'$ that is a set that we use for constructing this subtree.

Let us present \textsc{ConstructClassifiers} and \textsc{FormTree} procedures as Algorithm 1. The \textsc{FormTree} procedure does two steps. The first one \textsc{ChooseSplit} is choosing the test $B$ that is the choosing an attribute and the splitting by this attribute that maximize the objective function $\frac{G({\cal X}';B)}{P({\cal X}';B)}$. The result attribute index is $attr$ and the result split is $split$ variable. The second step \textsc{Divide} is the splitting processes itself. 
\begin{figure}[h!]
    \centering
    \includegraphics[width=400px]{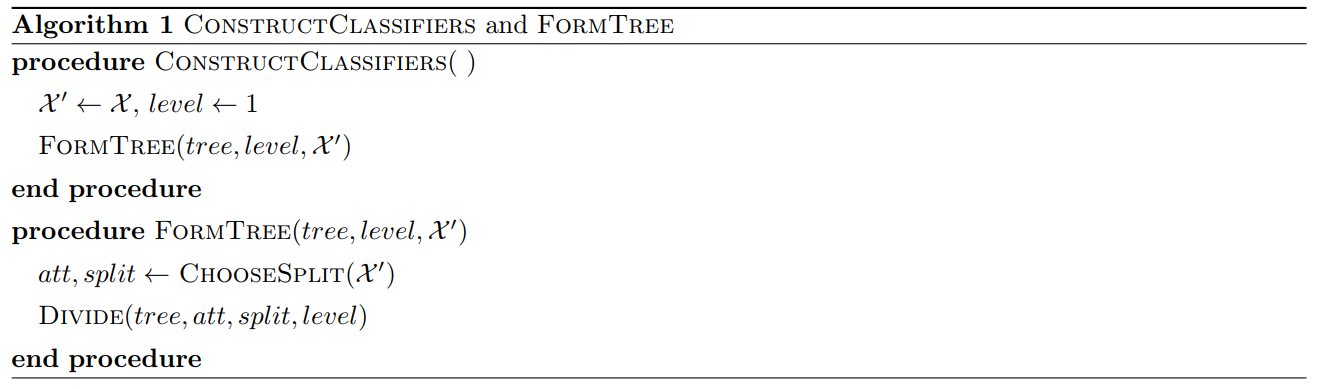}
    \end{figure}

Let us describe the \textsc{ChooseSplit} procedure that provides the best attribute index and split itself. It is presented in Algorithm 2. The procedure considers each attribute, and it has two different kinds of processing processes of the attribute depending on belonging to $RV$ or $DV$. Let us describe the procedure for a real-valued attribute $attr$. We have three steps.

 The first step is a sorting ${\cal X'}$ by $x_{attr}$ element of vectors. This procedure is $\textsc{Sort}({\cal X}',j)$. Assume that the result indexes in a sorted order are $(i_1,\dots,i_{z})$, where $z=|{\cal X}'|$. So, now we can split vectors ${\cal X'}$ by $\theta_{u}=(x_{attr}^{u}+x_{attr}^{u+1})/2$ and then there will be two sets ${\cal X}_1=\{X^{i_1},\dots,X^{i_u}\}$ and ${\cal X}_2=\{X^{i_{u+1}},\dots,X^{i_z}\}$, for $u\in\{1,\dots,z-1\}$.
 
 The second step is computing $pC_j[u]=|C_j^u|$, $pI[u]=I(\{X^{i_1},\dots,X^{i_u}\})$ and $pbI[u]=I(\{X^{i_u},\dots,X^{i_z}\})$, where $u\in\{1,\dots,z\}$, $C_j^u=\{w:w\in\{1,\dots,u\},y_{i_w}=j\}$, $j\in\{1,\dots,M\}$. We use the following formula for the values:
 
$pC_j[u]=pC_j[u-1]+1$ if $y_{i_u}=j$; and $pC_j[u]=pC_j[u-1]$ otherwise.

$pI[u]=pI[u-1]-\left(-\frac{pC_j[u-1]}{N}\log\frac{pC_j[u-1]}{N} + \frac{pC_j[u]}{N}\log\frac{pC_j[u]}{N}\right)$, if $y_{i_u}=j$. 

$pbI[u]=pbI[u+1]-\left(-\frac{pC_j[z]-pC_j[u]}{N}\log\frac{pC_j[z]-pC_j[u]}{N} + \frac{pC_j[z]-pC_j[u-1]}{N}\log\frac{pC_j[z]-pC_j[u-1]}{N}\right)$, if $y_{i_u}=j$. 


The third step is choosing maximum $\max\limits_{u\in\{1,\dots,z-1\}}\frac{G({\cal X}',u)}{P({\cal X}',u)}$, where $G({\cal X}',u)=I({\cal X}')-\frac{u}{N}\cdot pI[u] - \frac{N-u}{N}\cdot (pbI[u+1])$ and $P({\cal X}',u)=-\frac{u}{N}\cdot \log\left(\frac{u}{N}\right) -\frac{N-u}{N}\cdot \log\left(\frac{N-u}{N}\right)$. We use these formulas because they correspond to splitting ${\cal X}_1=\{X^{i_1},\dots, X^{i_u}\}$ and ${\cal X}_2=\{X^{i_{u+1}},\dots, X^{i_z}\}$, $I({\cal X}_1)=pI[u]$, $I({\cal X}_2)=pbI[u+1]$ and $I({\cal X}')=pI[z]$.

If we process a discrete-valued attribute from $DV$, then we can compute  the value of the object function when we split all elements of ${\cal X}'$ according value of the attribute. So ${\cal X}_w=\{i:X^i\in{\cal X}', x^i_{attr}=w\}$, for $w\in\{1,\dots, T_{attr}\}$.

Let us describe the processing of discrete-valued attributes. The first step is computing the case numbers of classes before split, the case numbers of classes after split, the case numbers for t values of current attribute.

The second step is calculating an entropy $I({\cal X})$ before split. The third step is calculating the entropies $I\left({\cal X}_{i}\right)$ after split to $t$ branches, information gain $G({\cal X}, B)=I({\cal X})-\sum_{i=1}^{t} \frac{\left|{\cal X}_{i}\right|}{|{\cal X}|} I\left({\cal X}_{i}\right)$ and potential information $P({\cal X}, B)=-\sum_{i=1}^{t} \frac{\left|{{\cal X}}_{i}\right|}{|{\cal X}|} \log \left(\frac{\left|{{\cal X}}_{i}\right|}{|{\cal X}|}\right)$.
The last step is calculating a gain ratio $\frac{G({\cal X};B)}{P({\cal X};B)}$.


Let us describe the \textsc{ChooseSplit} procedure that splits the set of vectors. The procedure also described in Algorithm 2  and 3. The \textsc{Divide} procedure recursively invokes the \textsc{FormTree} procedure for each set from sequence of sets $split$ for constructing child subtrees.
\begin{figure}[h!]
    \centering
    \includegraphics[width=400px]{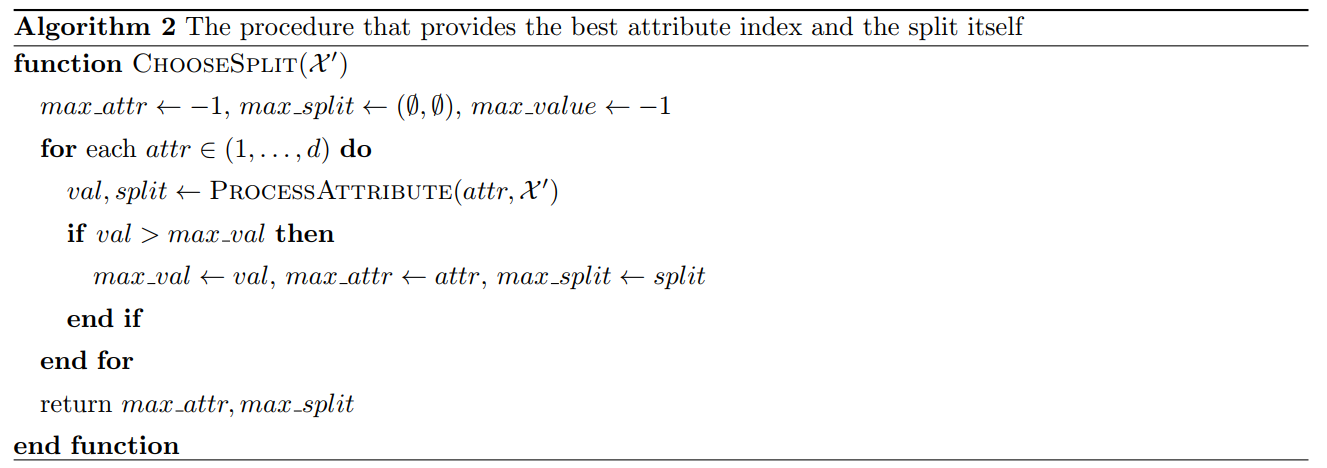}
    \end{figure}

\begin{figure}[h!]
    \centering
    \includegraphics[width=400px]{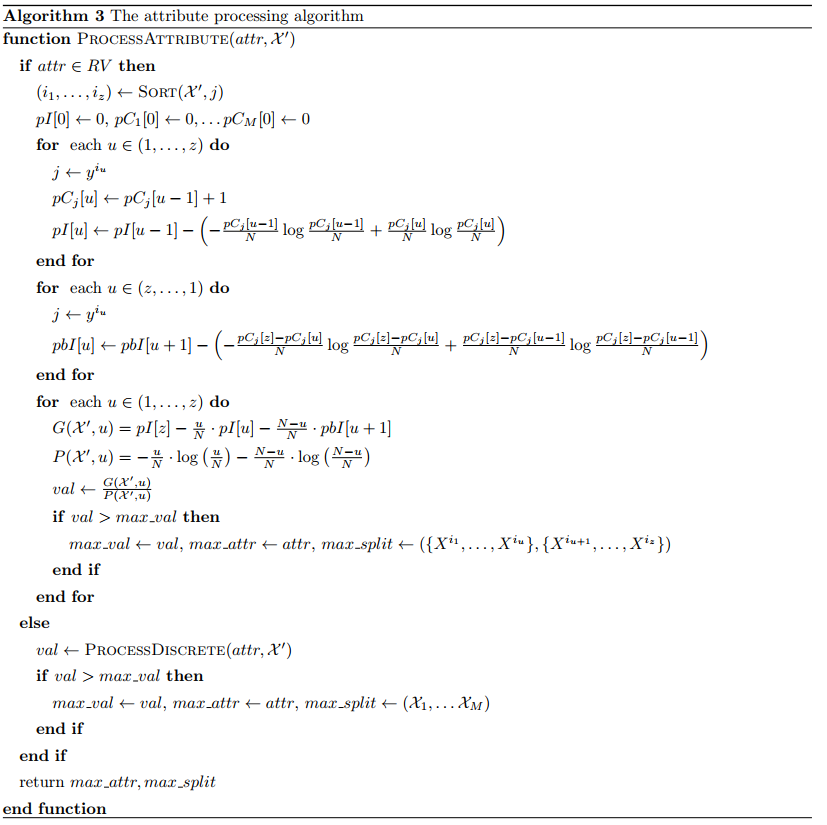}
    \end{figure}    
    
 \begin{figure}[h!]
    \centering
    \includegraphics[width=400px]{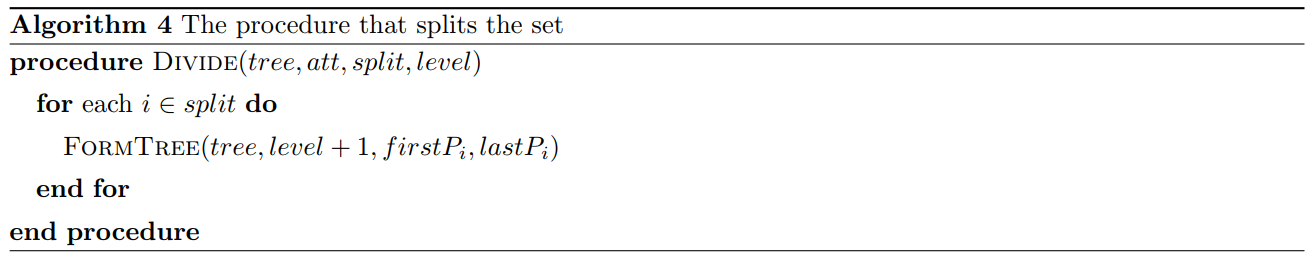}
    \end{figure}       
  \begin{figure}[h!]
    \centering
    \includegraphics[width=400px]{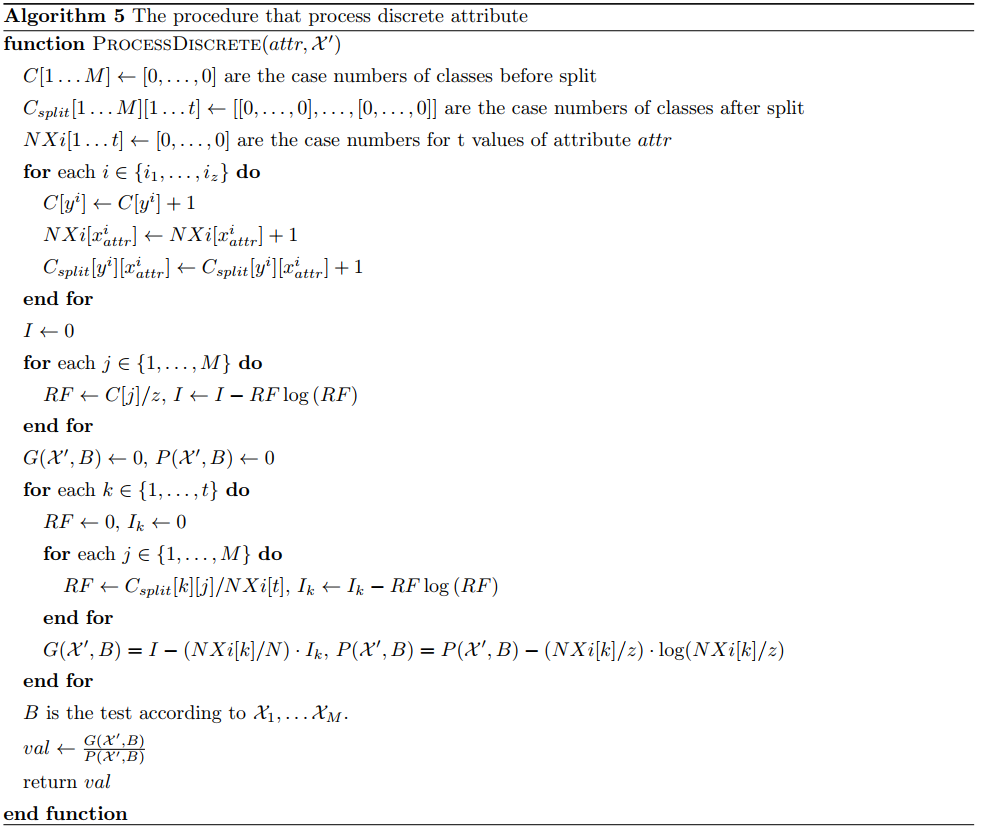}
    \end{figure}

 Let us discuss the running time of the algorithm.

\begin{theorem}\label{th:txc50}
The running time of C5.0 is $O(h\cdot d\cdot(M\cdot N+N \log N))$.
\end{theorem}
\Beginproof
In fact, $ChooseSplit$ takes the main time. That is why we focus on analyzing this procedure.
Let us consider a real-valued attribute. The running time for computing of $pI$, $pbI$ and $pC$ is $O(|{\cal X}'|)$. The running time for sorting procedure is $O(|{\cal X}'|\log |{\cal X}'|)$. The running time of computing a maximum of gain ratios for different splits is $O(|{\cal X}'|)$. Additionally, we should initialize $pC$ array that takes $O(M)$. The total complexity of this processing a real-valued attribute is $O(M+|{\cal X}'|\log |{\cal X}'|)$.

Let us consider a discrete-valued attribute. The cases processing time complexity is $O(N)$. An information gain $G({\cal X}, B)$ for some discrete attribute $B$ is calculated with $O(M \cdot t)$ running time, where $t$ is a number of attribute values, $M$ is a number of classes. An entropy before cutting $I({\cal X})$ is calculated with  $O(M)$ running time, an entropy after cutting is calculated in $O(M \cdot t)$. The potential information $P({\cal X}, B)$ is calculated with $O(t)$ running time. The gain ratio is calculated with  $O(1)$ running time.

Therefore the running time of processing of one discrete-valued attribute is $O(N + M \cdot t)$.

Note that if we consider all ${\cal X}'$ sets of one level of the decision tree, then we collect all elements of ${\cal X}$. Therefore, the total complexity for one level is $O(d\cdot(M\cdot N+N \log N))$, and the total complexity for the whole tree is $O(hd\cdot(M\cdot N+N \log N))$
\Endproof


\section{Improvement of the Classical C4.5/C5.0 algorithms}\label{sec:improvement}

\subsection{Improvement of Discrete-valued Attributes Processing}
If we process a discrete-valued attribute from $DV$, then we can compute  the value of the object function when we split all elements of ${\cal X}'$ according value of the attribute. So ${\cal X}_w=\{i:X^i\in{\cal X}', x^i_{attr}=w\}$, for $w\in\{1,\dots, T_{attr}\}$.

We will process all vectors of ${\cal X}'$  one by one.
Let us consider processing of current $u$-th vector $X^{i_u}$ such that $y^{i_u}=j$ and $x_{attr}^{i_u}=w$. Let us compute the following variables: $N_w$ is a number of elements of ${\cal X}_w$; $C_j$ is a number of vectors from ${\cal X}'$ that belongs to the class $j$; $C_{j,w}$ is a number of vectors from ${\cal X}_w$ that belongs to the class $j$; $P$ is a potential information; $I_w$ is $I({\cal X}_w)$; $I$ is information of ${\cal X}'$; $S=G({\cal X}',B)-I(X)$.
Assume that these variables contains values after processing $u$-th vector and $N_w',C_j', C_{j,w}',P',I'_w,I'$ and $S'$ contains values before processing $u$-th vector. The final values of the variables will be after processing all $z=|{\cal X}'|$ variables. We will recompute each variable according to the  formulas from Figure \ref{fig:formilas} (only variables that depends on $j$ and $w$ are changed)  
\begin{figure}[h]
    $N_w\gets N_w'-1$
    
    
    $C_j\gets C_j+1$
    
    $C_{j,w}\gets C_{j,w}'+1$
    
    $P\gets P'-\left(-\frac{N'_w}{z}\log\frac{N'_w}{z} + \frac{N_w}{z}\log\frac{N_w}{z}\right)$
    
    $I_w\gets I_w'-\left(-\frac{C_{j,w}'}{N_w'}\log\frac{C_{j,w}'}{N_w'} + \frac{C_{j,w}}{N_w}\log\frac{C_{j,w}}{N_w}\right)$
    
    $I\gets I'-\left(-\frac{C_{j}'}{z}\log\frac{C_{j}'}{z} + \frac{C_{j}}{z}\log\frac{C_{j}}{z}\right)$
    
    $S\gets S'-\left(-\frac{N_w'}{z}\log\frac{N_w'}{z} + \frac{N_w'}{z}\log\frac{N_w'}{z}\right)$
    \caption{Updating formulas}
    \label{fig:formilas}
\end{figure}

So, finally we obtain the \textsc{ProcessDiscrete} procedure from Algorithm 6.

   \begin{figure}[h!]
    \centering
    \includegraphics[width=400px]{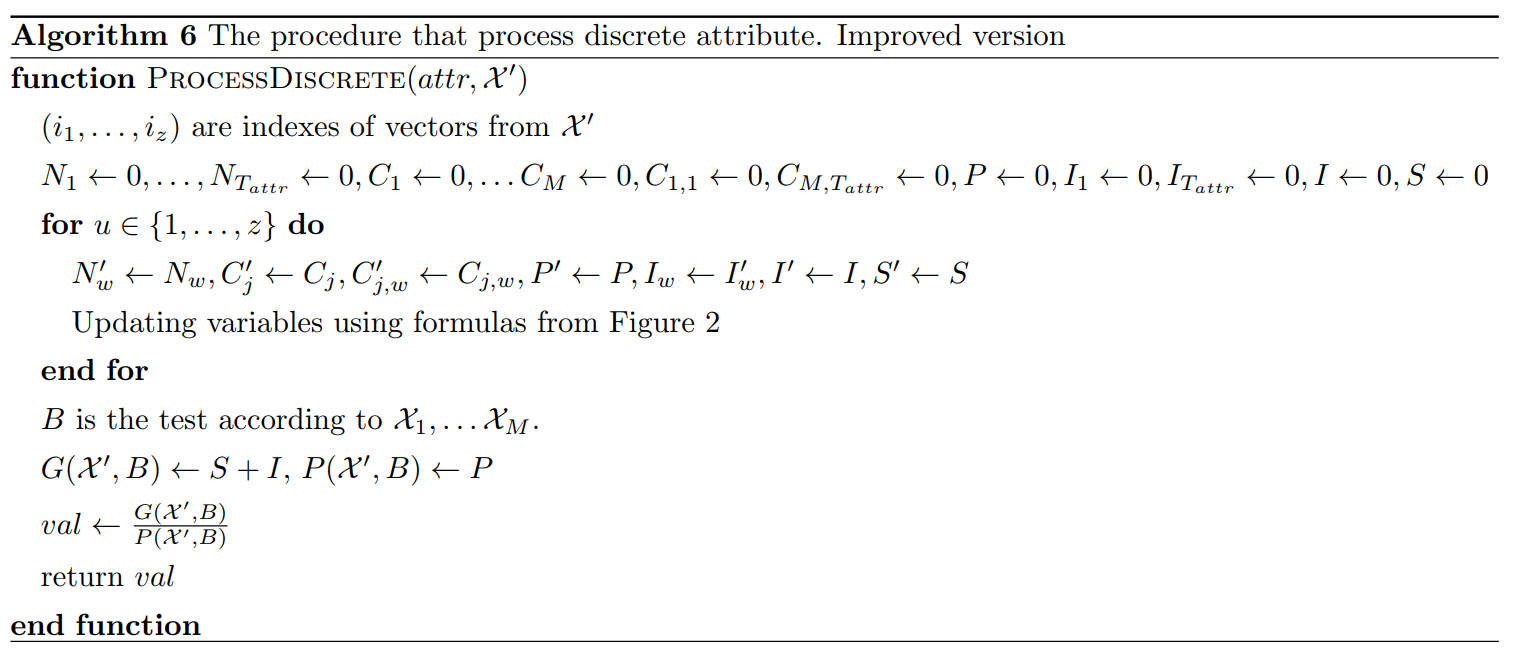}
    \end{figure}     
\subsection{Using a Self-balancing Binary Search Tree}
We suggest the following improvement. Let us use a self-balancing binary search tree \cite{cormen2001} data structure for $pC_j[u]$ and $C_{j,w}$.  As a self-balancing binary search tree we can use the AVL tree \cite{avl62,cormen2001} or the Red-Black tree \cite{g78,cormen2001}. This data structure can be used for   implementation of  mapping from set of indexes to set of values. 
We always mean that the data structure contains only indexes with a non-zero value, and other values are zero. We use indexes of non-zero elements as key for constructing the search tree and values as additional data that is stored in a corresponding node of the tree. In the paper we call this data structure as Tree Map. The data structure has three properties on running time. 
(i) Running time of adding, removing and inserting a new index (that is called key) to the data structure is $O(\log s)$, where $s$ is a number of keys in the tree or a number of indexes with non-zero values.
(ii) Running time of finding a value by index and modification of the value is $O(\log s)$ 
(iii)Running time of removing all indexes from the data structure and checking all indexes of data structure is $O(s)$, where $s$ is a number of indexes with non-zero values. 

If we use Tree Map, then we can provide the following running time.
\begin{lemma}
The running time of C5.0 that uses Tree Map (Self-balancing binary search tree) is $O(h\cdot d \cdot N \log N))$.
\end{lemma}
\Beginproof
Let us follow the proof of Theorem \ref{th:txc50}. If we do not need to initialize the  $pC_j[u]$ and $C_{j,w}$, but erase these values after processing an attribute, then this procedure takes $O(|{\cal X}'|)$ steps. So, the running time for processing a real-valued attribute becomes $O(N\log N + N \log N)=O(N \log N)$, and for a discrete-valued attribute, it is $O(N\log N)$ because we process each vector one by one and recompute variables that takes only $O(\log N)$ steps for updating values of $C_{j,w}$ and $O(1)$ steps for other actions. Therefore, the total complexity is $O(hd\cdot N \log N)$. 
\Endproof

\section{Quantum C5.0}\label{sec:qauntumalgo}

The key idea of the improved version of C5.0 algorithm is using the D{\"u}rr and H{\o}yer's algorithm for maximum search and Amplitude Amplification algorithm.

These two algorithms in combination has the following property:

\begin{lemma}
Suppose, we have a function $f:\{1,\dots,K\}\to \mathbb{R}$ such that the running time of computing $f(x)$ is $T(K)$. Then, there is a quantum algorithm that finds argument $x_0$ of maximal $f(x_0)$, the expected running time of the algorithm is $O(\sqrt{K}\cdot T(K))$ and the success probability is at least $\frac{1}{2}$.
\end{lemma}
\Beginproof
For prove we can use D{\"u}rr and H{\o}yer's for minimum search \cite{dh96} with replacing Grover's Search algorithm by Amplitude amplification version for computing $f(x)$ from \cite{bhmt2002}.   
\Endproof

Using this Lemma we can replace the maximum search by attribute in \textsc{ChooseSplit} function and use \textsc{ProcessAttribute} as function $f$. Let us call the function \textsc{QChooseSplit}. Additionally, for reducing an error probability, we can repeat the maximum finding process $\log d$ times and choose the best solution. The procedure is presented in Algorithm 7.

   \begin{figure}[h!]
    \centering
    \includegraphics[width=400px]{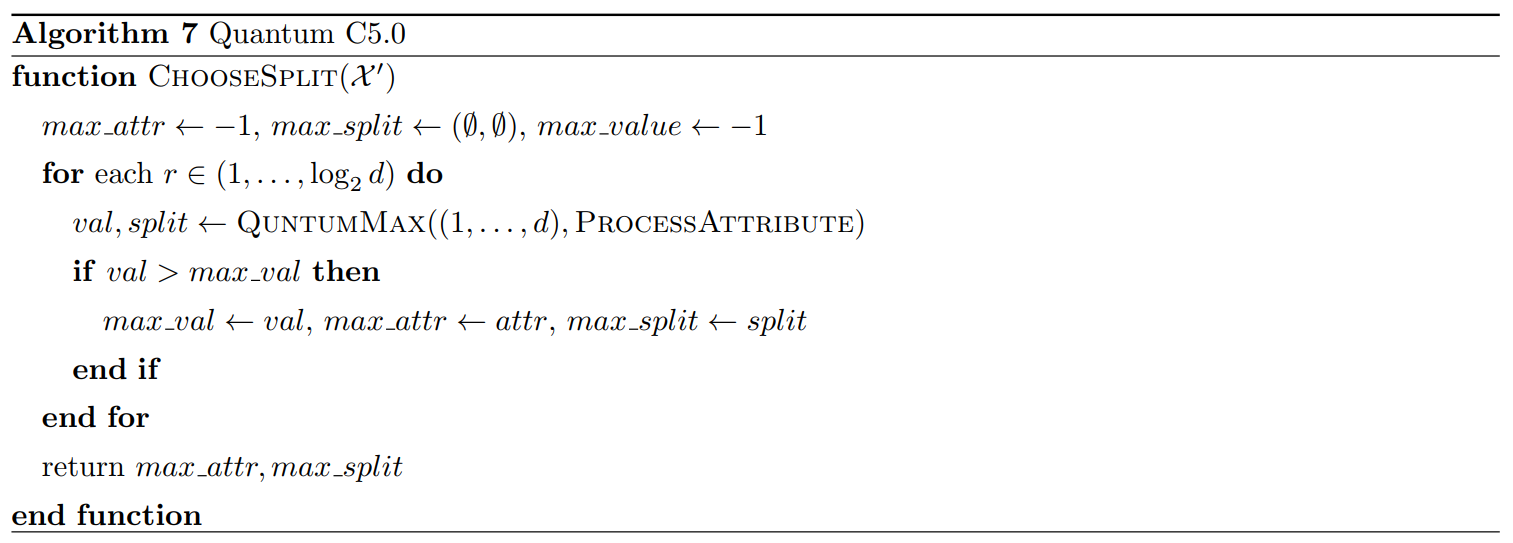}
    \end{figure}

\begin{theorem}
The running time of the Quantum C5.0 algorithm is $O\big((h\sqrt{d}N \log N) \log d\big)$. The success  probability of QC5.0 is $O\big((1-\frac{1}{d})^k\big)$, where $k$ is a number of inner nodes (not leaves).
\end{theorem}
\Beginproof
The running time of \textsc{ProcessAttribute} is $O(|{\cal X}'|\log |{\cal X}'|)$. So the running time of maximum searching is $O(\sqrt{d}|{\cal X}'|\log |{\cal X}'|)$. With repeating the algorithm, the running time is $O(\sqrt{d}|{\cal X}'|\log |{\cal X}'|\log d)$. If we sum the running time for all nodes, then we obtain $O\big(h\sqrt{d}N \log N) \log d\big)$.
The success probability of the D{\"u}rr and H{\o}yer's algorithm is $\frac{1}{2}$. We call it $\log d$ times and choose a maximum among $\log d$ values of gain ratios. Then, we find a correct attribute for one node with a success probability  $O\big(1-\frac{1}{2^{\log d}}\big)=O\big(1-\frac{1}{d}\big)$. We should find correct attributes for all nodes except leaves. Thus, the success probability for the whole tree is equal to $O\big((1-\frac{1}{d})^k\big)$, where $k$ is a number of internal nodes (not leaves).
\Endproof

\section{Conclusion}
Firstly, we have suggested a version of the C4.5/C5.0 algorithm with Tree Map (Self-balancing binary search tree, for example Read-Black tree or AVL tree) data structure. This version has a better running time. Secondly, we have presented a quantum version of the C5.0 algorithm for classification problem. This algorithm demonstrates almost quadratic speed-up with respect to a number of attributes.


\bibliographystyle{alpha}
\bibliography{tcs}

\begin{thebibliography}{BHMT02}

\bibitem[AdW17]{aw2017}
Srinivasan Arunachalam and Ronald de~Wolf.
\newblock Guest column: a survey of quantum learning theory.
\newblock {\em ACM SIGACT News}, 48(2):41--67, 2017.

\bibitem[Amb17]{a2017}
A.~Ambainis.
\newblock Understanding quantum algorithms via query complexity.
\newblock {\em arXiv:1712.06349}, 2017.

\bibitem[AVL62]{avl62}
George~M Adel'son-Vel'skii and Evgenii~Mikhailovich Landis.
\newblock An algorithm for organization of information.
\newblock In {\em Doklady Akademii Nauk}, volume 146, pages 263--266. Russian
  Academy of Sciences, 1962.

\bibitem[BHMT02]{bhmt2002}
G.~Brassard, P.~H{\o}yer, M.~Mosca, and A.~Tapp.
\newblock Quantum amplitude amplification and estimation.
\newblock {\em Contemporary Mathematics}, 305:53--74, 2002.

\bibitem[c5019]{c50main}
C5.0: An informal tutorial, 2019.
\newblock url={https://www.rulequest.com/see5-unix.html}.

\bibitem[CLRS01]{cormen2001}
T.~H Cormen, C.~E Leiserson, R.~L Rivest, and C.~Stein.
\newblock {\em Introduction to Algorithms-Second Edition}.
\newblock McGraw-Hill, 2001.

\bibitem[DH96]{dh96}
Christoph Durr and Peter H{\o}yer.
\newblock A quantum algorithm for finding the minimum.
\newblock {\em arXiv preprint quant-ph/9607014}, 1996.

\bibitem[DW01]{dw2001}
Ronald De~Wolf.
\newblock {\em Quantum computing and communication complexity}.
\newblock 2001.

\bibitem[Eth10]{ml1}
Alpaydin Ethem.
\newblock Introduction to machine learning.
\newblock 2010.

\bibitem[Fri99]{fri}
J.~H. Friedman.
\newblock Greedy function approximation: A gradient boosting machine.
\newblock 1999.

\bibitem[Gro96]{g96}
Lov~K Grover.
\newblock A fast quantum mechanical algorithm for database search.
\newblock In {\em Proceedings of the twenty-eighth annual ACM symposium on
  Theory of computing}, pages 212--219. ACM, 1996.

\bibitem[GS78]{g78}
L.~J Guibas and R.~Sedgewick.
\newblock A dichromatic framework for balanced trees.
\newblock In {\em Proceedings of SFCS 1978)}, pages 8--21. IEEE, 1978.

\bibitem[HTF09]{GTB}
T.~Hastie, R.~Tibshirani, and J.~H. Friedman.
\newblock {\em The Elements of Statistical Learning:Data Mining, Inference, and
  Prediction. Second Edition}.
\newblock 2009.

\bibitem[Hun]{h62}
EB~Hunt.
\newblock Concept learning: An information processing problem. 1962.

\bibitem[Jor]{quantumzoo}
Stephen Jordan.
\newblock Bounded error quantum algorithms zoo.
\newblock https://math.nist.gov/quantum/zoo.

\bibitem[KKS19]{kks2019}
K.~Khadiev, D.~Kravchenko, and D.~Serov.
\newblock On the quantum and classical complexity of solving subtraction games.
\newblock In {\em Proceedings of CSR 2019}, volume 11532 of {\em LNCS}, pages
  228--236. 2019.

\bibitem[Kop18]{ko2018}
Dawid Kopczyk.
\newblock Quantum machine learning for data scientists.
\newblock {\em arXiv preprint arXiv:1804.10068}, 2018.

\bibitem[KQ02]{DT}
R.~Kohavi and J.~R. Quinlan.
\newblock Data mining tasks and methods: Classification: decision-tree
  discovery.
\newblock {\em Handbook of data mining and knowledge discovery. – Oxford
  University Press}, 2002.

\bibitem[KS19]{ks2019}
K.~Khadiev and L.~Safina.
\newblock Quantum algorithm for dynamic programming approach for dags.
  applications for zhegalkin polynomial evaluation and some problems on dags.
\newblock In {\em Proceedings of UCNC 2019}, volume 4362 of {\em LNCS}, pages
  150--163. 2019.

\bibitem[LHAJ84]{cart}
Breiman L., Friedman~J. H., Olshen~R. A., and ~Stone~C. J.
\newblock {\em Classification and regression trees}.
\newblock 1984.

\bibitem[NC10]{nc2010}
Michael~A Nielsen and Isaac~L Chuang.
\newblock {\em Quantum computation and quantum information}.
\newblock Cambridge university press, 2010.

\bibitem[Qui79]{q79}
J~R. Quinlan.
\newblock Discovering rules by induction from large collections of examples.
\newblock {\em Expert systems in the micro electronics age}, 1979.

\bibitem[Qui86]{ml2}
J.~R. Quinlan.
\newblock Induction of decision trees.
\newblock {\em Machine learning}, pages 81--106, 1986.

\bibitem[Qui87]{q87}
J.~R. Quinlan.
\newblock Simplifying decision trees.
\newblock {\em International journal of man-machine studies}, 27(3):221--234,
  1987.

\bibitem[Qui96]{c45}
J.~R. Quinlan.
\newblock Improved use of continuous attributes in c4.5.
\newblock {\em Journal of Artificial Intelligence Research}, pages 77--90,
  1996.

\bibitem[RJ15]{c50}
Pandya R. and Pandya J.
\newblock C5. 0 algorithm to improved decision tree with feature selection and
  reduced error pruning.
\newblock {\em International Journal of Computer Applications.}, pages 18--21,
  2015.

\bibitem[SSP14]{ssp2014qnn}
Maria Schuld, Ilya Sinayskiy, and Francesco Petruccione.
\newblock The quest for a quantum neural network.
\newblock {\em Quantum Information Processing}, 13(11):2567--2586, 2014.

\bibitem[SSP15]{ssp2015}
Maria Schuld, Ilya Sinayskiy, and Francesco Petruccione.
\newblock An introduction to quantum machine learning.
\newblock {\em Contemporary Physics}, 56(2):172--185, 2015.

\end{thebibliography}

\end{document}